\documentclass[letterpaper, 10 pt, conference]{ieeeconf}  
\IEEEoverridecommandlockouts                              
\usepackage{amsmath, amssymb}
\usepackage{amsfonts, amsthm}
\usepackage{cite}
\usepackage{color}
\usepackage{xcolor}
\usepackage{bm}
\usepackage{url}
\usepackage{algorithm}
\usepackage{algpseudocode}
\usepackage{svg}
\usepackage{caption}
\usepackage{subcaption}
\newcommand{\lly}[1]{{\color{blue}{LLY:#1}}} 
\usepackage{verbatim}
\def\BibTeX{{\rm B\kern-.05em{\sc i\kern-.025em b}\kern-.08em
    T\kern-.1667em\lower.7ex\hbox{E}\kern-.125emX}}
\newcommand{\lf}[1]{{\color{purple}{LF:#1}}}

\usepackage{makecell}

\title{\LARGE \bf
DART: A Compact Platform for Autonomous Driving Research${}^{*}$
}

\author{Lorenzo Lyons$^{1}$, Thijs Niesten$^{1}$, Laura Ferranti$^{1}$
\thanks{*This research is supported by the NWO-TTW Veni project
HARMONIA (no. 18165).}
\thanks{$^{1}$The authors are with the Reliable Robot Control Lab, Department of Cognitive Robotics,
        Delft University of Technology, 2628 CD Delft, The Netherlands
        {\tt\small \{l.lyons,T.E.J.Niesten,l.ferranti\}@tudelft.nl}}%
}

\begin{document}

\maketitle
\thispagestyle{empty}
\pagestyle{empty}

\maketitle

\begin{abstract}
This paper presents the design of a research platform for autonomous driving applications, the Delft’s Autonomous-driving Robotic Testbed (DART). Our goal was to design a small-scale car-like robot equipped with all the hardware needed for on-board navigation and control while keeping it cost-effective and easy to replicate. To develop DART, we built on an existing off-the-shelf model and augmented its sensor suite to improve its capabilities for control and motion planning tasks. We detail the hardware setup and the system identification challenges to derive the vehicle's models. Furthermore, we present some use cases where we used DART to test different motion planning applications to show the versatility of the platform. Finally, we provide a git repository with all the details to replicate DART, complete with a simulation environment and the data used for system identification.
\end{abstract}

\section{Introduction}
Autonomous driving technologies have seen significant advancements in recent years leading to many companies around the world investing in, testing, and, in some cases, commercializing autonomous driving services, such as robotaxis \cite{Web_article_robotaxis_worldwide} and truck platooning \cite{Web_article_paltooning_companies}. Despite these encouraging signals, the day when fully autonomous cars will be a large percentage of traffic participants is decades ahead of us \cite{Web_article_realistic_prospects}, and much research is sill needed to unlock the full benefits of autonomous driving. 

To deploy these technologies in our cities, extensive testing and validation of tailored navigation algorithms are necessary steps to understand potential limitations, but also gain users' trust. Assessing these methods directly on full-scale platforms, however, entails for significant costs and long time scales, not to mention safety and regulatory restrictions. 

Driving simulators such as \cite{carla,khusro2020} are a cost effective alternative and allow to build complex urban environments that would be otherwise prohibitive to replicate in real life. Yet, even with the recent improvement in simulation accuracy, no simulator is able to fully capture the interactions among multiple road users and also among the different components of an autonomous vehicle. The latter is indeed a complex mechatronics system where physical phenomena such as road-tire interactions, actuator and sensor dynamics, electronic and software components are strictly connected and affect each other.  

Small-scale testing platforms provide an intermediate step (or a compromise) between a purely simulated system and a full-scale autonomous car. The cost of developing a 1/10 scaled-down platform is orders of magnitude smaller (if we also consider the cost of experiments) than doing so for a full scale vehicle. Furthermore, being a real system, it can provide significant added value over purely simulated results in understanding how the developed method performs in hardware. Despite the advantages of using small-scale platforms in autonomous driving research, very few models are available on the market, most of which focus on machine-learning perception algorithms for racing, making them challenging to use for control and multi-agent applications. As a result many research groups have developed their own customized platform, leading to reproducibility and accessiblity issues. 

\begin{figure}[t]
    \centering
    \includegraphics[width=0.99\linewidth]{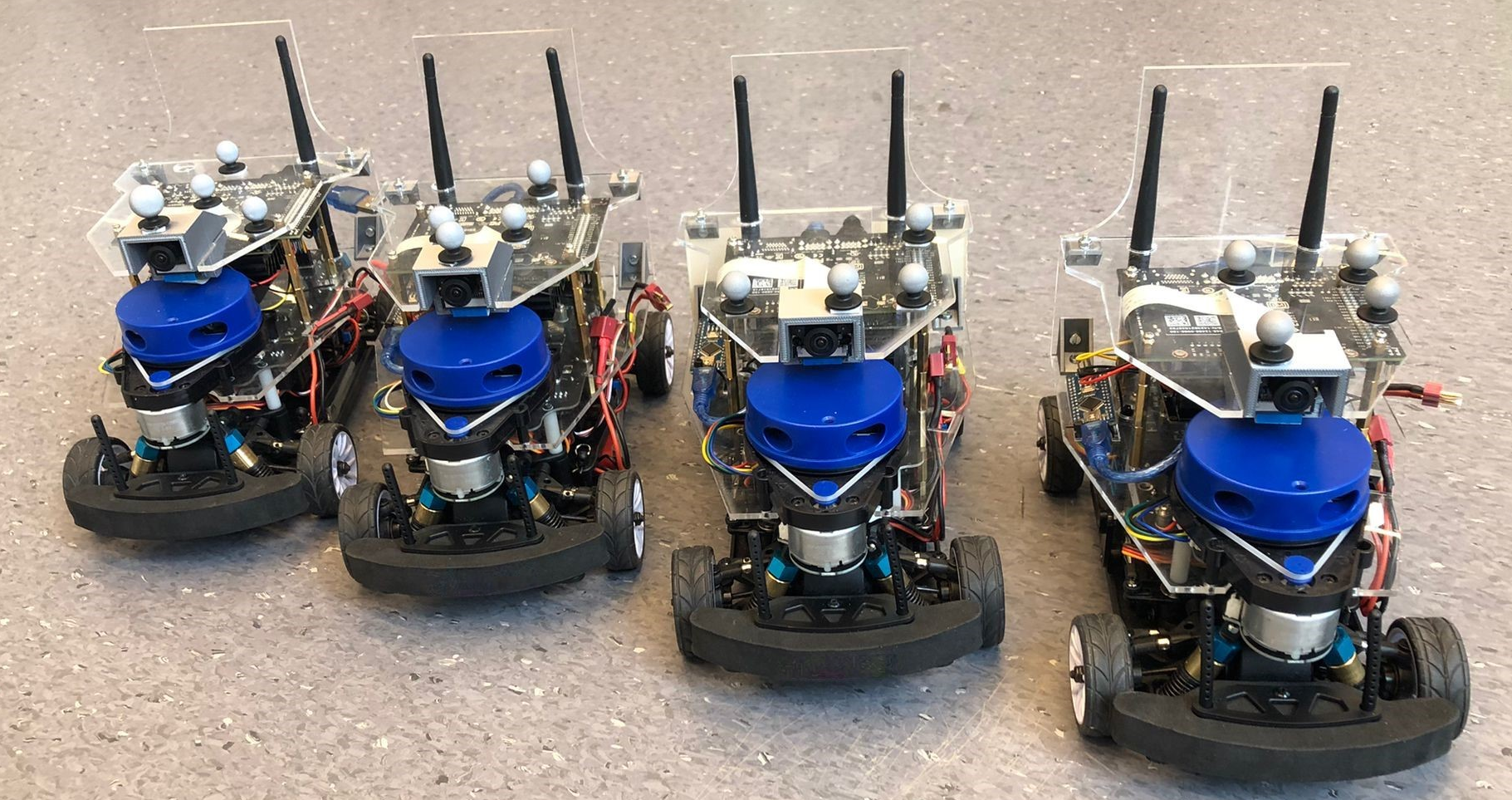}
    \caption{DART, a small-scale robotic platform for autonomous driving research.}
    \label{fig:jetracers_picture}
\end{figure}

With the previously mentioned issues in mind this paper presents the Delft's Autonomous-driving Robotic Testbed (DART), shown in Figure~\ref{fig:jetracers_picture}. This affordable and easy to build small-scale car-like robot is designed for both single and multi-vehicle autonomous driving experiments.

\section{related works}
This section reviews small-scale car-like robots, roughly between 1:5 and 1:20 of a full vehicle. Full-scale research platforms and differential wheeled robots like TurtleBot~\cite{turtlebot} and Duckiebot~\cite{duckiebot} are considered out of scope for the present paper and will thus not be covered. We will describe the most well-known platforms starting from largest (1:5 scale) to smallest (1:16 scale) highlighting their main features, summarised in table \ref{tab:related_works_table} for convenience. Notice that the listed prices are indicative, as costs may vary over time and across different countries. 

The AutoRally \cite{AUTORALLY} is 1:5 scale vehicle-like robot platform developed for aggressive autonomous outdoor driving at the Georgia Tech Autonomous Racing Facility. The AutoRally platform is based on a 1/5 scale RC trophy truck. Due to its size, it can include a powerful mini desktop computer for state-of-the-art computationally heavy algorithms. The mini desktop can communicate  through 2.4GHz/5GHz Dual-Band High-Speed WiFi and through an extra added 900 MHz XBee Pro providing a low latency, low-bandwidth wireless communication channel. Featuring high-end computation units, sensors, and actuators the total cost of this platform is around $5000\$$. To add these components to the main body and make them ready for outdoor conditions, several parts are customized.

The MIT RACECAR \cite{MIT_RACECAR} (Rapid Autonomous Complex-Environment Competing Ackermann-steering Robot) is a 1:10 scale vehicle-like robot platform developed by MIT. The RACECAR uses a Traxxis Rally Car platform with a powerful Nvidia Jetson TX1 computer to process all the data from the attached sensors. Communication is provided by 2.4GHz/5GHz Dual-Band High-Speed WiFi supported by the Nvidia Jetson TX1. Featuring high-end computation units, sensors, and actuators the total cost of this platform, including the lidar and Jetson TX1 computer, is $\approx 3700\$$. The system is also equipped with a self-made Vedder ESC that can prove difficult to replicate. 

The BARC \cite{BARC} (Berkeley Autonomous Race Car) is a 1:10 scale car-like robotic platform developed by UC Berkeley. The latest version, V4.0, consists of a Tamiya chassis and a Nvidia Jetson Nano, that supports the same Communication features as the more powerful TX1 computer. The platform, featuring an front facing Intel Depth camera and rear facing RGB camera, costs around $950\$$. 

The MuSHR \cite{MUSHR} (Multi-agent System for non-Holonomic Racing) is a 1:10 scale platform developed at the University of Washington’s Paul G. Allen School of Computer Science Engineering. MuSHR can be built using the Redcat Racing Blackout SC 1:10 chassis and has two additional features compared to the BARC, which are a 2D Lidar and a bumper switch for collision detections. This platform comes at a cost of around $1050\$$.

The Donkey Car \cite{DONKEY_CAR} is a 3D printed platform that can easily be attached to a 1:16 scale vehicle-like robot platform, with a few mentioned in the build instructions \cite{DONKEY_CAR} but they are no longer available. The computing module on the Donkey Car is a Raspberry Pi that uses 2.4Ghz WiFi for communication. This platform features only one RGB camera and costs around $300\$$.  

\begin{table*}[t]
    \centering
    \begin{tabular}{|c|r|c|c|c|c|}
    \hline
          \textbf{Platform}  & \textbf{Cost [\$]} & \textbf{Scale} & \textbf{Computing} & \textbf{Sensors} & \textbf{Drawbacks} \\ \hline
        AutoRally~\cite{AUTORALLY}  & 5.000,00  & 1:5 & Powerfull custom-build computer & \makecell{Lord Microstrain 3DM-GX4-25 IMU \\ Emisphere P307 GPS  \\ 2 RGB cameras \\ Odemetry sensor} & \makecell{High number of custom parts} \\ \hline
        MIT RACECAR~\cite{MIT_RACECAR}  & 3.663,00  & 1:10 & Nvidia Jetson TX1 &  \makecell{Hokuyo UST-10LX \\ Stereolabs ZED \\ Structure.io depth camera} & \makecell{Self made ESC\\No shipping outside USA} \\ \hline
        BARC~\cite{BARC}  & 950,00  & 1:10 & Jetson Nano & \makecell{Intel Realsense D435i \\ RGB camera} &  \\ \hline
        MuSHR~\cite{MUSHR}  & 1050,00  & 1:10 & Jetson Nano & \makecell{Intel Realsense D435i \\ YDlidar X4 \\ VEX bumber switch} &  \\ \hline
        Donkey Car~\cite{DONKEY_CAR}  & 300,00  & 1:16 & Raspberry Pi & RGB camera & \makecell{Low computational power \\ One camera sensor} \\ \hline
    \end{tabular}
    \caption{Main features of the small-scale car-like platforms available in the literature.}
    \label{tab:related_works_table}
\end{table*}

\textit{Contributions.} The main factor that hinders the widespread adoption of the small-scale platforms available in literature is the need for custom parts. Indeed many prestigious research institutions have a team of skilled technicians specifically dedicated to design and maintain the robotic platforms, yet this may not be the case for other research labs. Some DIY robots, like the Donky Car project, solve this problem by relying on easily printable 3D parts, yet they feature a limited computation power and sensor-actuator suite, diminishing their appeal as research platforms. DART aims to fill this gap in the literature by presenting itself as a low-cost yet versatile platform. The contributions of this paper are threefold:
\begin{itemize}
    \item DART. We present the result of the design process behind the platform, as well as a list of parts and indications on the level of skill required to build it.  
    
    \item System identification. A reliable dynamic model is essential to accurately control the vehicle, yet in contrast to full-scale cars where a vast amount of literature is available, much fewer works focus on model-fitting practices for small-scale cars. In this paper we present a model identification process specifically designed for 1:10 scale car-like robots. 

    \item Hardware and software setup. To ease reproducibility and speed up the setup of the platform the building instructions and the code relative to low level control, system identification and some examples of high level controllers, as well as a simulation environment can be found in the GitHub repository \cite{git_repo}. 
    
\end{itemize}

\section{DART}
This section presents the main features of the proposed robotic platform. For a quick overview of the functionalities and the relative required parts see Table \ref{tab:component list}, while for a complete hardware and setup guide please refer to the GitHub repository \cite{git_repo}.

DART's design adhers to the following criteria: 
\noindent\subsubsection*{Accessibility} the total cost of the platform should be as low as possible, making it affordable for most research institutions, even for educational purposes. 
\noindent\subsubsection*{Reproducibility} the platform should be based on a commercially available hardware and the custom parts should be as few as possible. \noindent\subsubsection*{Versatility} the platform should lend itself to a broad variety of research fields, thus maximizing the number of researchers using the same platform, ultimately facilitating knowledge transfer. 

With these criteria in mind, the robot is based on the commercially available JetracerPro AI kit \cite{jetracer}. This platform is originally designed for machine learning tasks and features a good computation unit (the Nvidia Jetson Nano), high reachable speeds due to the powerful brushed electric motor and 4-wheel drive. 

Our goal is to use the platform to test navigation and control algorithms for autonomous and cooperative driving. Hence, to be able to use such a vehicle for these applications, we augmented the base kit as follows. We introduced a custom-made magnetic and infrared wheel encoder and an IMU necessary to produce odometry measurements, as well as an Arduino to process the raw sensor data. We also added a Lidar needed for localization and mapping (note that a camera with fish-eye lenses is already available in the base kit). Furthermore, after some testing, we upgraded the servomotor used for steering and included an external LiPo battery to power both the latter and the longitudinal driving electric motor. These two upgrades have proven to significantly improve the consistency of the measured longitudinal acceleration and steering angle, largely increasing control performances. The total cost of the platform is around €700 and requires little technical knowledge to assemble. 

One challenge we faced was how to optimize the available space to accommodate the extra hardware given the size of the platform without compromising its driving performance. For this, we developed custom PVC plates to accommodate the augmented sensor suite and the encoder to measure wheel velocity. The PVC plates can be produced from the .stl files available in the GitHub repository \cite{git_repo} and many companies or university workshops offer online laser cutting services. The encoder requires fitting small magnets on the gear of the main shaft, yet no soldering is required and this operation requires low technical skills.

\begin{table*}[t]
\centering
\def\arraystretch{1.2}
\begin{tabular}{|c|c|c|r|}
\hline
\textbf{Component}           & \textbf{Functionality}                         & \textbf{Notes}                                         & \textbf{Cost {[}€{]}} \\ \hline
JetracerPro AI kit  & Base robotic platform                 & 18650 batteries are not included              & 420,00       \\ \hline
YLidar X4 Lidar            & Localization and mapping              & -                                             & 100,00          \\ \hline
Encoder             & Wheel velocity measurement            & Requires medium-low technical skills to build & 10,00           \\ \hline
IMU BNO055          & Acceleration and yaw rate measurement & -                                             & 40,00           \\ \hline
Arduino Nano        & Collect data from encoder and IMU     & Price includes jumper-wires and adaptors      & 30,00           \\ \hline
LiPo battery        & Power servo and driving motor         & Price is for 2 batteries and a charger        & 30,00           \\ \hline
Servomotor DS3225   & Improved steering consistency         & -                                             & 20,00           \\ \hline
PVC mounting plates & Accommodate augmented sensor suite     & Needs to be custom made                       & 20,00           \\ \hline
                    &                                       & \textbf{Total cost}                                    & 670,00          \\ \hline
\end{tabular}
\caption{Overview of the parts needed to build DART.}
\label{tab:component list}
\end{table*}
\section{System identification}\label{sec:system identification}
Building reliable vehicle models is not only required to run model-based controllers such as MPC, but is also necessary to develop realistic simulators to test any kind of controller before deploying it on the real hardware. 

A dynamic system, such as a robot, can be represented by a system of ordinary differential equations $\dot{\hat{\bm{x}}}=\hat{\bm{f}}(\hat{\bm{x}},\bm{u})$, where $\hat{\bm{x}}\in \mathbb{R}^{n_x},\bm{u}\in\mathbb{R}^{n_u}$ are the real state and control input, respectively. System identification refers to the problem of building a function $\bm{f}$, i.e., the model of the system, that is able to adequately approximate the real dynamics $\hat{\bm{f}}$~\cite{system_id}. This can be achieved by recording measurements of the state and control action, collected in a matrix $\bm{X}$ of size $N\times M$, where $N$ is the number of data points and $M$ is the dimension of the input data. Note that the field of control the term "input" usually refers to the control action, while in system identification and machine learning the same term indicates the inputs to the model. In the remainder of this section we will follow the latter's terminology, i.e., $M=n_x+n_u$. For each row in $\bm{X}$, the corresponding measured system output is collected in a matrix $\bm{Y}$ of size $N\times n_x$. In this paper we will make use of parametric models, that is, we assume that the function $\bm{f}$ has a fixed structure where a set of parameters $\bm{p}\in \mathbb{R}^{n_p}$ can be chosen in order to better approximate the real dynamics, i.e., $\bm{f}=\bm{f}(\bm{X},\bm{p})$. The optimal set of parameters $\bm{p}$ is chosen as the solution the following minimization problem:
\begin{align}\label{eq:fitting intro}
\min_{\bm{p}}& \sum^N_{k=0} \mathcal{D}\left(\bm{f}\left(\bm{X}_k,\bm{p}\right),\bm{Y}_k\right),
\end{align}
where $\bm{X}_k$ and $\bm{Y}_k$ indicate the $k$-th row in the repsective matrices, and $\mathcal{D}$ is a function that measures the distance between the observed data point $\bm{Y}_k$ and the model output $\bm{f}\left(\bm{X}_k,\bm{p}\right)$. A typical choice for $\mathcal{D}$ is the squared error, i.e. $\mathcal{D}(\bm{f}\left(\bm{X}_k,\bm{p}\right),\bm{Y}_k)=(\bm{f}\left(\bm{X}_k,\bm{p}\right)-\bm{Y}_k)^{\top} (\bm{f}\left(\bm{X}_k,\bm{p}\right)-\bm{Y}_k)$. 

The most critical aspect of parametric model identification is the design of the function $\bm{f}(\bm{X},\bm{p})$ and the definition of reasonable parameter bounds, since this will heavily influence the ability of the model to correctly represent the data. While for full-scale vehicles there is a rich literature featuring various models~\cite{rajamani2011vehicle}, empirical formulas like the Pacejka tire model~\cite{pacejka1992magic} for tire forces estimation, best practices for data collection and reference values for the fitting parameters~\cite{lateral_dynamics_practices}, much less is available for small-scale cars. Furthermore, the dynamic models used for full-scale cars do not easily transfer to their small-scale counterparts, since for example the latter do not feature pressurised tires and the weight scale is around 2-3 orders of magnitude different. As a result, the range of typical values for some important parameters (e.g., the cornering stiffness) can be significantly different between full-scale and small-scale cars, making model identification particularly challenging for the latter. 

In the remaining part of this section we will describe how to obtain reliable kinematic and dynamic bicycle models~\cite{kin_dyn_bicycle_models_borrelli} for these small-scale vehicles. We will describe how to obtain $\left\{\bm{X},\, \bm{Y}\right\}$ in problem~\eqref{eq:fitting intro} from the raw sensor data and how to define reasonable parameter bounds. The values of the parameters are obtained by numerically solving problem~\eqref{eq:fitting intro} where $\mathcal{D}$ has been chosen as the square error, using the ADAM gradient descent based algorithm implemented in PyTorch. Table~\ref{tab:fitting parameter values} shows the obtained parameter values. Notice that all the raw data and the code for data processing and model fitting is available in the GitHub repository~\cite{git_repo}, as well as a simulation environment that uses the obtained vehicle models. 

\subsection{Kinematic bicycle model}
The kinematic bicycle model consists of the following system of ordinary differential equations:
\begin{align}\label{eq: kin model}\begin{bmatrix}\dot{x}\\\dot{y}\\\dot{\eta}\\\dot{v}\end{bmatrix}&=\begin{bmatrix}v\cos{\eta}\\v\sin{\eta}\\v \tan(\delta)/l\\(F_m + F_f)/m\end{bmatrix}
\end{align}
Where the states $x$, $y$, $\eta$ and $v$ are the position of the rear axle, orientation, and longitudinal velocity of the vehicle, respectively. The mass and distance between the front and rear axles of the robot are indicated by $m$ and $l$, respectively. $F_m$ and $F_f$ are the motor and friction forces, respectively. For full-scale cars, the motor characteristic curve that indicates the longitudinal force $F_m$ transmitted to the wheels, and the steering angle $\delta$ as a function of the steering input are usually provided by the manufacturer. For small-scale cars, on the other hand, throttle $\tau$ and steering input $s$ are provided to an ESC module and a servomotor, respectively, both of which accept normalized non-dimensional values between $[-1,1]$. Identifying how these inputs are related to motor torque and steering angles represents an additional complication in the model fitting process. 

The kinematic bicycle model~\eqref{eq: kin model} for small-scale cars requires the identification of the following sub-models: $F_f(v)$, $F_m(\tau,v)$, $\delta(s)$, as well as the actuation delays. Simply collecting driving data and fitting the model poses significant challenges since the fitting function for each sub-model needs to be user designed, and the relative parameters need to be initialized to some reasonable value in order to numerically solve Problem~\eqref{eq:fitting intro}. For this reason our approach was to isolate each component and progressively build the full model. Attempting to identify one sub-model at a time allows us to tailor the type of test (e.g., step input or sinusoidal input test) to better highlight its contribution, while clearly visualizing the measured data aids in the design of the fitting function. 

\subsubsection{Longitudinal dynamics}
\begin{figure}[t]
	\centering
	\includegraphics[width=\linewidth]{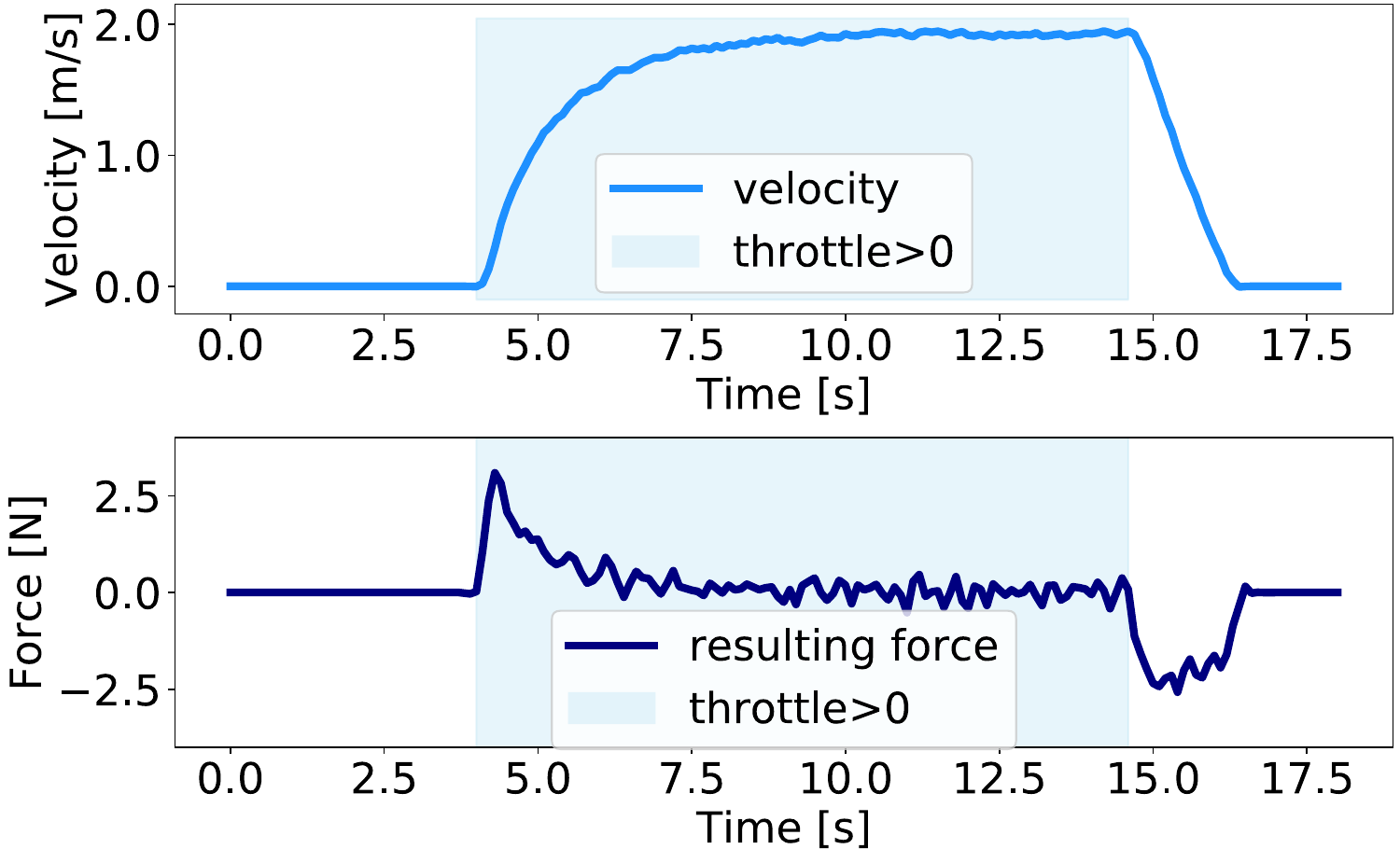}
	\caption{The velocity profile (top) and the estimated resulting force acting on the vehicle (bottom) measured in response to a step throttle input (shaded area).}\label{fig:step response}
\end{figure}

We performed a series of step response tests for different throttle values on a smooth and level surface, while the steering is kept equals to zero. The collected raw data is a time series of throttle-velocity pairs. The acceleration is then obtained by numerically deriving the velocity, while the resulting longitudinal force $F=F_m+F_f$ acting on the vehicle is estimated by multiplying the acceleration by the mass of the vehicle $m=1.67$ Kg that can be easily measured. The training inputs $\bm{X}$ are thus the throttle and velocity, while the training labels ${\bm{Y}}$ are the resulting measured longitudinal force. Figure~\ref{fig:step response} shows an example of the step response data.

\begin{figure}[t]
	\centering
	\includegraphics[width=\linewidth]{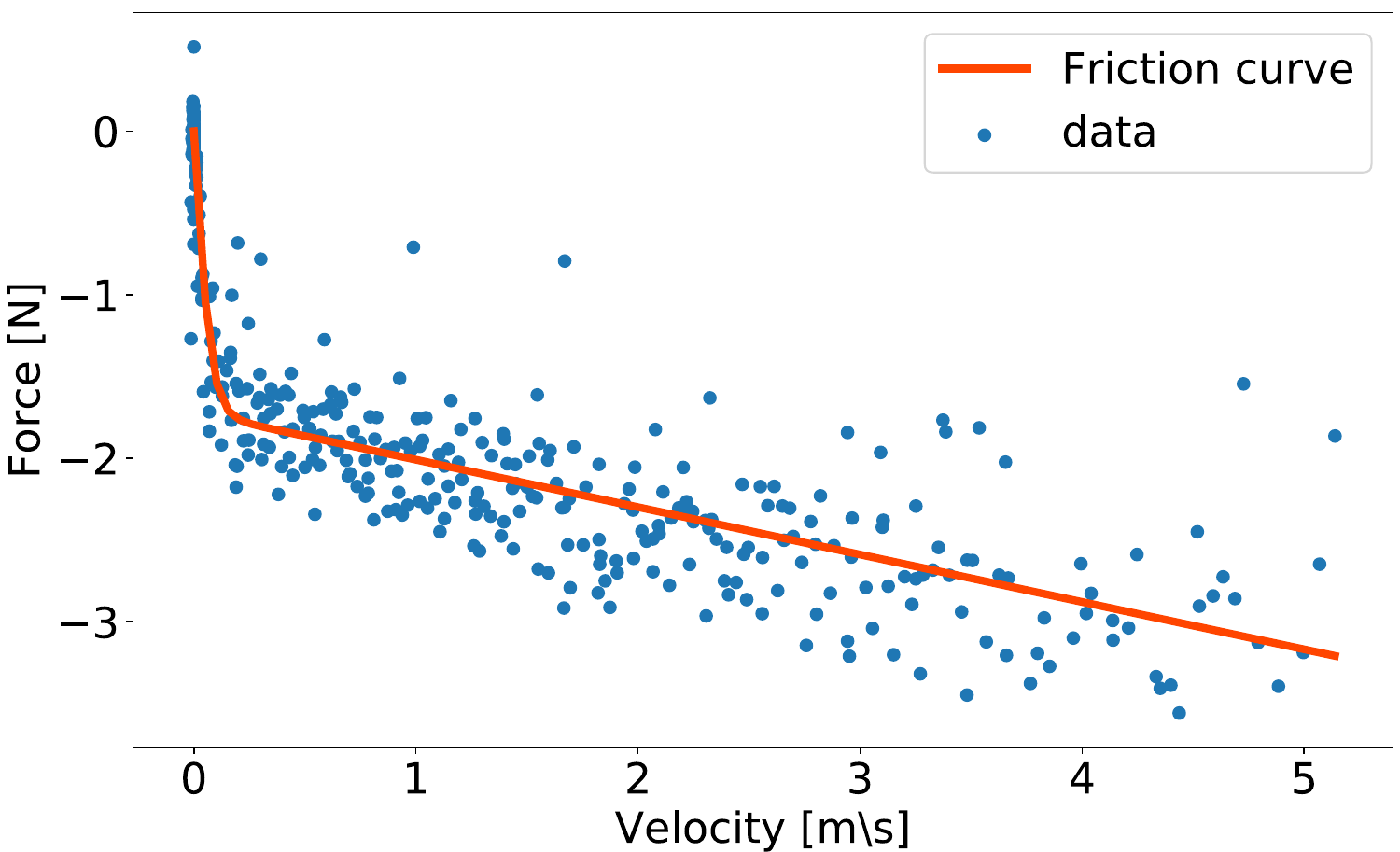}
	\caption{Friction curve fitting results.}\label{fig:friction curve}
\end{figure}

We start by modeling the friction force $F_f$. To isolate its contribution we selected the data with $\tau = 0$, that is, when the motor is switched off and vehicle is slowing down due to friction alone. By using the data showed in Figure~\ref{fig:friction curve}, we designed the friction curve as:
\begin{align*} 
    F_f(v)= - (a \tanh(b v) + v c)
\end{align*}
Where $a,b,c$ are the fitting parameters. Notice that reasonable parameter initialization can be found by plotting the resulting friction curve over the data.

\begin{figure}[t]
	\centering
	\includegraphics[width=\linewidth]{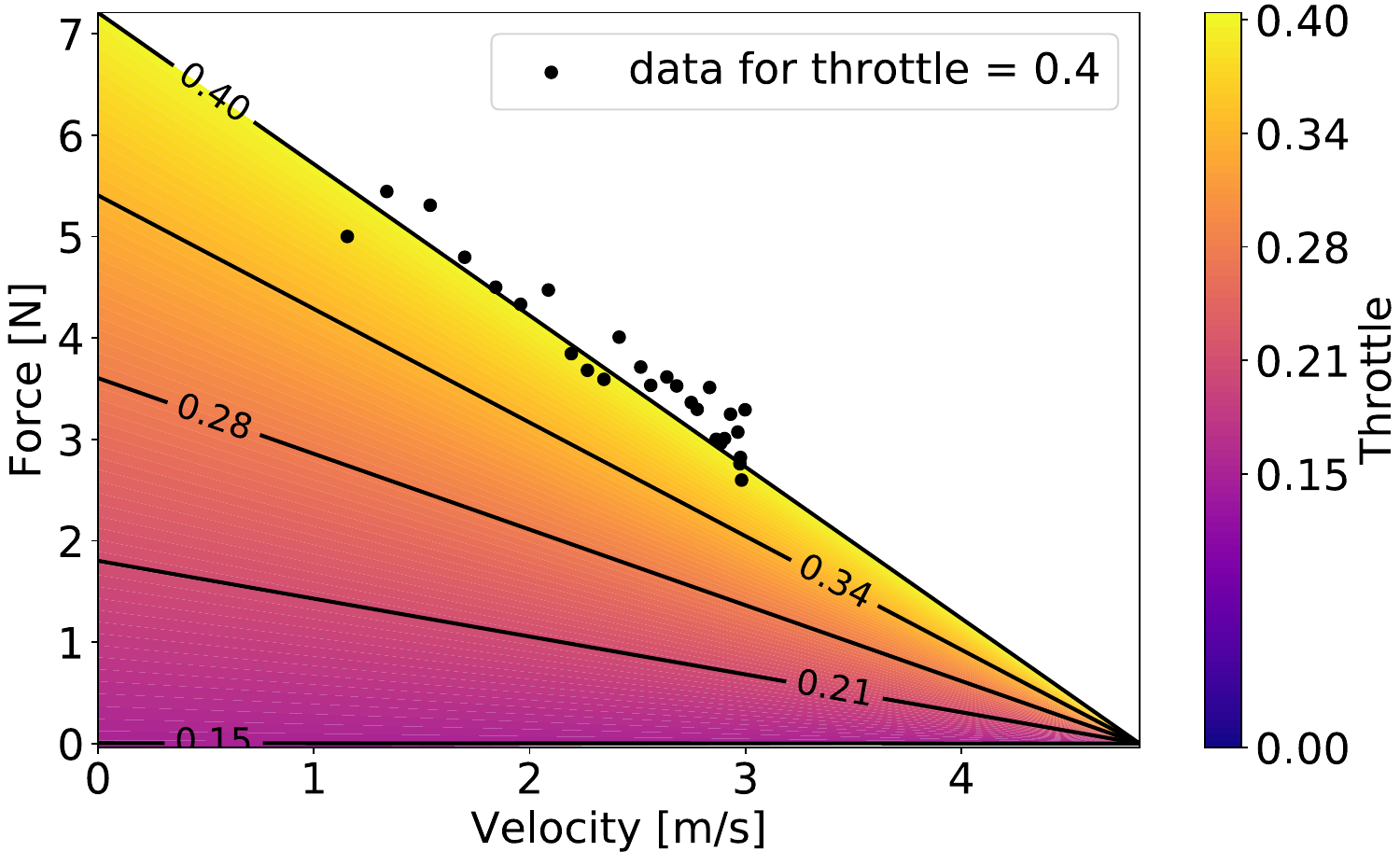}
	\caption{Motor curve fitting results.}\label{fig:motor curve}
\end{figure}

We now procede to model the motor force $F_m$. Since the friction term $F_f$ has been successfully identified, we can define new training labels $\bm{Y}$ as the estimated motor force $F_m$, obtained as $\bm{Y}=F-F_f(\bm{X})$. We designed $F_m$ in accordance to the characteristic curve of a brushed electric motor \cite{brushed_dc_motors}:
\begin{align*}
F_m &= (d - v e)\Tilde{\tau}\\
\Tilde{\tau} &= (\tau+g) 0.5 (\tanh(100(\tau+g))+1)
\end{align*}
Where $d,e,g$ are the fitting parameters. Note that the term $\Tilde{\tau}$ is an approximation of the function $\Tilde{\tau} =\max(0,\tau+g)$ but is continuously differentiable. This is highly desirable if the model will be used for model-based control such as MPC. Reasonable first guess parameter values for $d$ and $e$ can be obtained by plotting the measured step response data for a certain throttle value on the $\{v$,$F_m\}$, plane. The initial value for $g$ was estimated by increasing the throttle from $\tau=0$ until the vehicle begins moving. Figure~\ref{fig:motor curve} shows the step response data for $\tau=0.4$ and $1\,\leq v \leq 3\,\textrm{m/sec}$ used to initialize $d$ and $e$, and the overall characteristic motor curve, note that the latter was obtained using the full dataset featuring $\tau$ in the range $\tau\in[0.15,0.4]$.

\subsubsection{Steering input to steering angle mapping}
\begin{figure}[t]
	\centering
	\includegraphics[width=\linewidth]{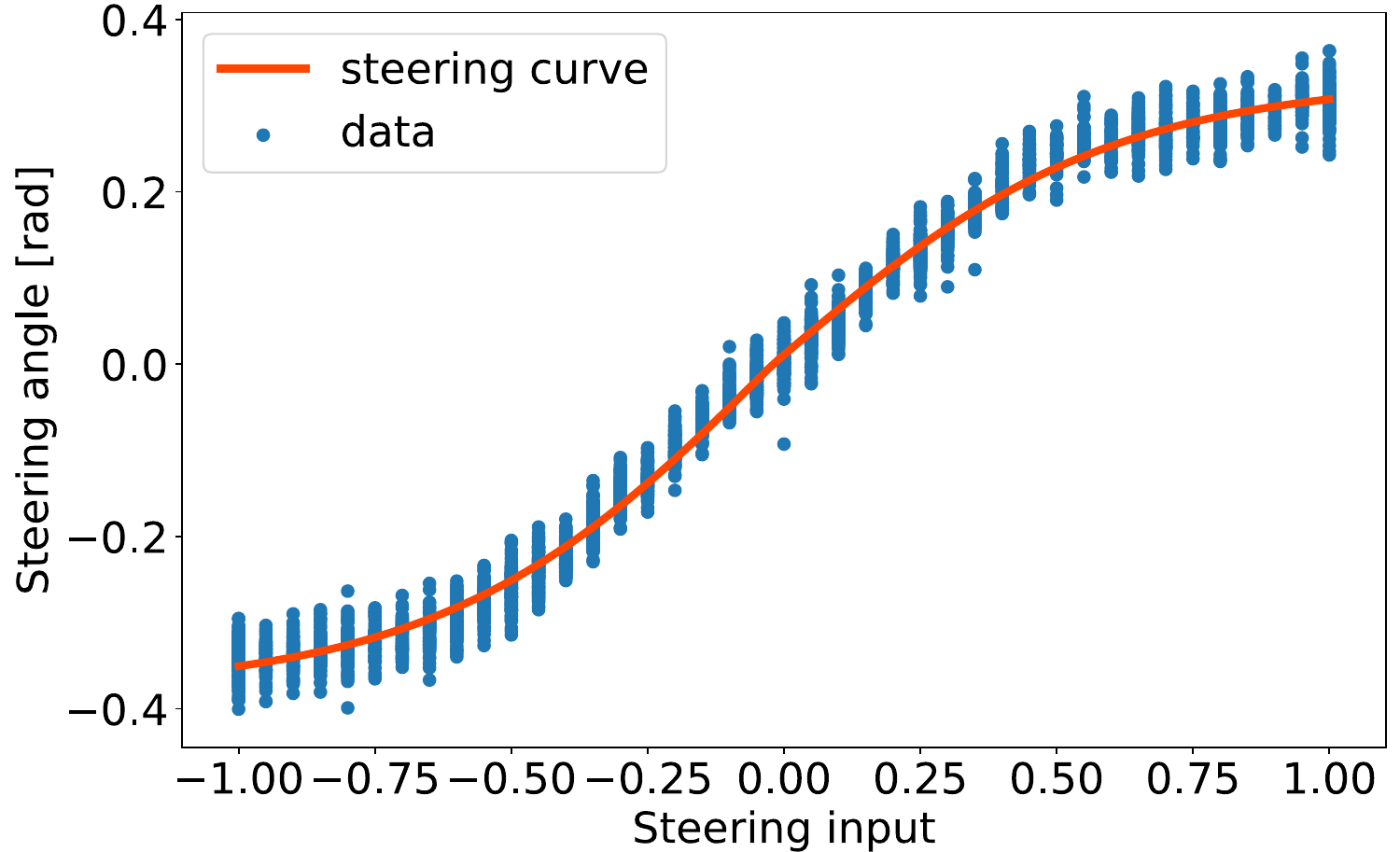}
	\caption{Steering input to steering angle static mapping.}\label{fig:steering map}
\end{figure}

The servomotor used for steering only accepts steering inputs in the range $s\in[-1,1]$, we therefore need to identify how they relate to actual steering angles $\delta$. To do so, we preformed a series of constant steering input tests while keeping the throttle at a constant value. The raw data consists of the resulting yaw rate $\dot{\eta}$ and the vehicle longitudinal speed $v$. By inverting the $\eta$ dynamics (i.e., the third equation in~\eqref{eq: kin model}) we are able to estimate the steering angle as: 
\begin{align}\label{eq: delta inverse kin model}
    \delta = \arctan\left(\frac{l\dot{\eta} }{v}\right)
\end{align}
We can thus create a dataset using the steering input $s$ as the training input $\bm{X}$ and the measured steering angle $\delta$ as the training labels $\bm{Y}$. Using the data shown in Figure~\ref{fig:steering map} we define the steering input to steering angle mapping as the weighted sum of two sigmoidal curves. This is needed to capture the asymmetry between steering to the left and steering to the right. The curve $\delta(s)$ is defined as:

\begin{align}\label{eq:steering map}
    \delta(s) &=\Tilde{w} \Tilde{a} \tanh(\Tilde{b} (s + \Tilde{c}))+(1-\Tilde{w})\Tilde{d} \tanh(\Tilde{e} (s + \Tilde{c}))\\
    \Tilde{w} &= 0.5 (\tanh(30(s+\Tilde{c}))+1)
\end{align}
Reasonable first guess values were identified empirically from the shape of the measured data.

\subsubsection{Actuation delays} The longitudinal actuation delay was measured by lifting the robot off the ground (so to reduce the system's inertia) and measuring the time delay between a throttle step input and a change in the wheel speed $v$. From these tests the longitudinal actuation delay was found to be negligible (in the order of $0.01\,\text{sec}$). The steering delay was instead obtained by providing the vehicle with a low frequency sinusoidal steering input. This allowed us to capture the reaction time of the steering, that is, the time interval before seeing a reaction after sending a steering command. The delay was then estimated by performing the cross correlation~\cite{cross_correlation} between the steering angle input $\delta(s)$ and the measured steering angle obtained from equation~\eqref{eq: delta inverse kin model}. The steering delay was found to be around $0.15\,\text{sec}$ and is thus not negligible.


\subsection{Dynamic bicycle model}
DART allows one to perform highly dynamic maneuvers (e.g., racing). In such a context, a kinematic model is no longer a good representation of the vehicle dynamics, since it neglects the lateral dynamics of the vehicle. Indeed it relies on the assumption that no lateral slipping occurs, that is, there is no lateral motion in the vehicle body frame. This assumption holds until the tires are able to provide enough lateral friction force to counter the centrifugal force during curves. The centrifugal force can be evaluated as $F_y = m\, \dot{\eta}\, v$. For high speed and high yaw rate, the assumption that lateral slip is negligible is no longer valid and the kinematic bicycle model fails. The dynamic bicycle model, on the other hand, does account for the lateral dynamics and relies on a tire model to predict tire force saturation, providing better model accuracy at high speeds.

The dynamic bicycle model is defined as:
\begin{align}\label{eq: dyn bike}
    \begin{bmatrix}\dot{x}\\\dot{y}\\\dot{\eta}\\\dot{v}_x\\\dot{v}_y\\\dot{\omega}\end{bmatrix}&=
    \begin{bmatrix}v_x\cos{\eta}- v_y\sin{\eta}\\
    v_x\sin{\eta}+ v_y\cos{\eta}\\
    \omega\\
    (\Hat{F}_{x,r}+\Hat{F}_{x,f}\cos{\delta}-\Hat{F}_{y,f}\sin{\delta})/m +\omega v_y\\
    (\Hat{F}_{y,r}+\Hat{F}_{y,f}\cos{\delta}+\Hat{F}_{x,f}\sin{\delta})/m -\omega vx\\
    (l_f(\Hat{F}_{y,f}\cos{\delta}+\Hat{F}_{x,f}\sin{\delta})-l_r\Hat{F}_{y,r})/I_z\end{bmatrix},
\end{align}
where $v_x$ and $v_y$ are the velocity components of the centre of mass measured in the vehicle body frame. $l_f$ and $l_r$ are the distances between the centre of mass and the front and rear axle, respectively. $\omega$ is the yaw rate and $I_z$ is the moment of inertia around the vertical axes. The front and rear tire forces components $\Hat{F}_{f,x}$, $\Hat{F}_{f,y}$, $\Hat{F}_{r,x}$ and $\Hat{F}_{r,y}$ are measured in the respective tire's body frame. The evaluation of the tire forces is the most critical component of the dynamic bicycle model and a vast amount of literature is available for full-scale cars. One of the most famous tire models is the Pacejka magic formula \cite{pacejka1992magic}, defined as:
\begin{align}\label{eq: pacejka tire}
\Hat{F}_y = D \sin (C \arctan (B \alpha - E (B \alpha - \arctan (B \alpha))))
\end{align}
where $\alpha$ is the slip angle, defined as the angle from the direction of motion of the tire and the tire axis. Front and rear slip angles are defined as:
\begin{align}
    \alpha_f &= -\arctan(v_y + \omega lf) + \delta \label{eq: slip front}\\
    \alpha_r &= -\arctan(v_y - \omega lr). \label{eq: slip rear}
\end{align}
To identify the parameters $D,C,B,E$ we collected driving data keeping a constant steering angle and gradually increasing the longitudinal velocity. Since we need to measure both longitudinal and lateral velocities in the vehicle body frame we used an external motion capture system, as the on-board sensors are not able to measure such quantities. The raw data thus consists of the vehicle's centre of mass position and orientation in the global reference frame. We then computed the time derivatives to measure the absolute velocity $\Tilde{v}_x$, $\Tilde{v}_y$ and acceleration $\dot{\Tilde{v}}_x$, $\dot{\Tilde{v}}_y$, as well as the yaw rate $\omega$ and it's time derivative $\dot{\omega}$. The training inputs $\bm{X}$ are thus the front and rear slip angles evaluated as in equations~\eqref{eq: slip front} and~\eqref{eq: slip rear}, where the velocities in the vehicle body frame are evaluated as:
\begin{align}
    \begin{bmatrix}v_x\\v_y\end{bmatrix} =& 
    \begin{bmatrix}\cos{(-\eta)}&-\sin{(-\eta)}\\
                    \sin{(-\eta)}&\cos{(-\eta)} \end{bmatrix}
    \begin{bmatrix}\Tilde{v}_x\\\Tilde{v}_y\end{bmatrix}
\end{align}
The training labels $\bm{Y}$ are the lateral forces in the tire body frame and have been evaluated by solving the equations of motion of the body in the absolute frame of reference:
\begin{align}
    \begin{bmatrix}
        F_x\\
        F_{y,f}\\
        F_{y,r}
    \end{bmatrix}=
    \begin{bmatrix}
        \cos{\eta}&-\sin{\eta}&-\sin{\eta}\\
        \sin{\eta}&\cos{\eta}&\cos{\eta}\\
        0&l_f&-l_r
    \end{bmatrix}^{-1}
    \begin{bmatrix}
        \dot{\Tilde{v}}_x / m\\
        \dot{\Tilde{v}}_y / m\\
        \dot{\omega} / I_z
    \end{bmatrix}
\end{align}

\begin{figure}[t]
	\centering
	\includegraphics[width=\linewidth]{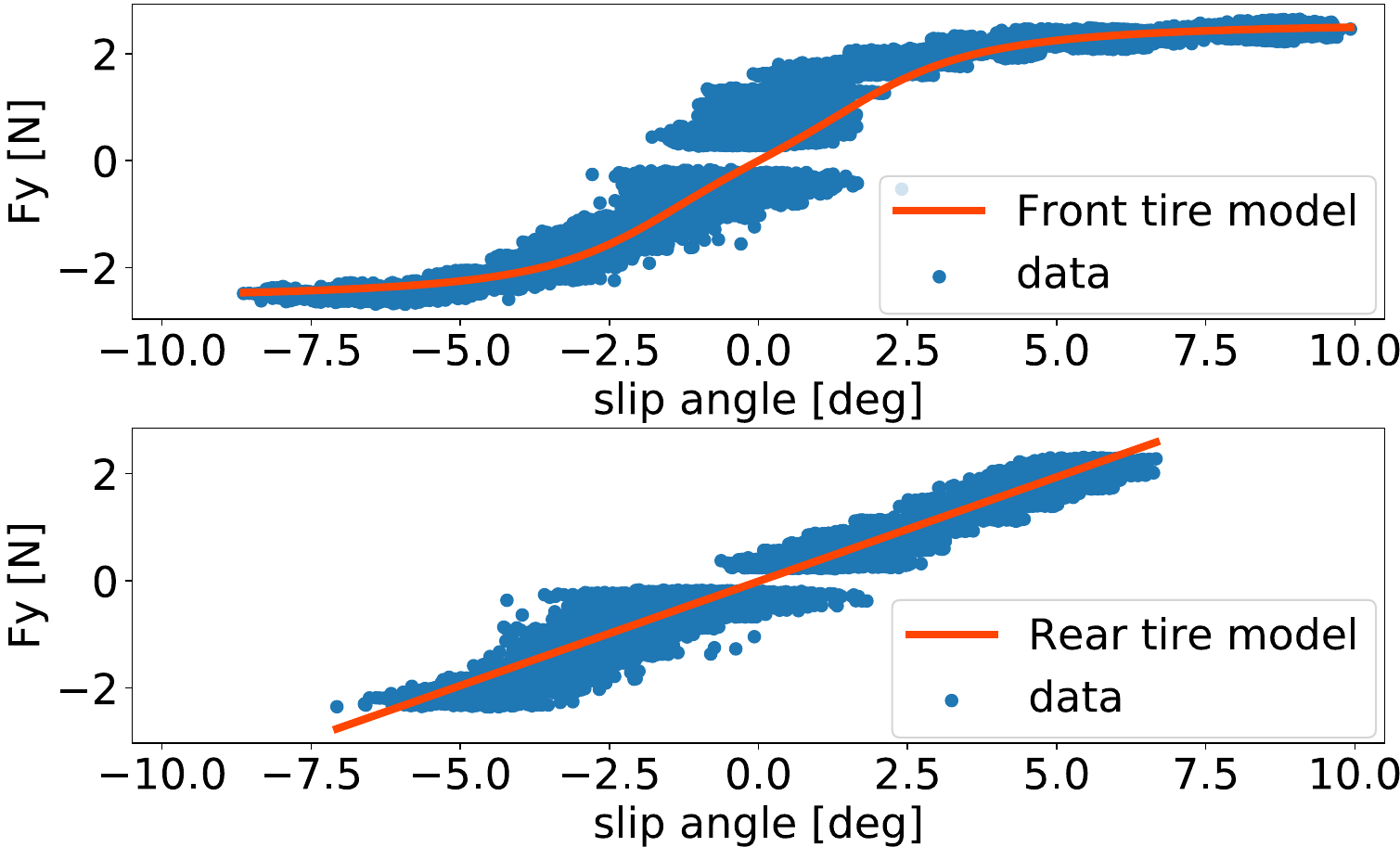}
	\caption{Lateral tire force model for the front tire (top) and rear tire (bottom).}\label{fig:tire_models}
\end{figure}

Where $F_x$ is the total longitudinal force in the vehicle frame, $F_{y,f}$ and $F_{y,r}$ are the front and lateral forces in vehicle frame, and $I_z\approx 0.006513 \text{Kgm}^2$ was estimated considering the robot as a rectangle with uniformly distributed mass of size $l \times w$, where $w=0.1\text{m}$ is the width of the vehicle. The lateral forces in the tire frame of reference are finally evaluated as: 
\begin{align}
    \Hat{F}_{y,f}&= \sin{(-\delta) F_{x,f} + \cos{(-\delta)} F_{y,f}}\\
    \Hat{F}_{y,r}&= F_{y,r}
\end{align}
 Where $F_{x,f}=\frac{1}{2}F_{x}$ since the vehicle features a 4 wheel drive and we assume that the motor force is equally shared by front and rear axle. Figure~\ref{fig:tire_models} shows the obtained data and fitting results. Notice that since the rear tire does not reach saturation we chose to model it with a simple linear relation instead of equation~\eqref{eq: pacejka tire}, i.e., as:
 \begin{align}
     F_{y,r} = C_r \alpha_r
 \end{align}
\begin{table}[t]
\centering
\begin{tabular}{|c|c|c|c|c|c|c|c|}
\hline
\rule{0pt}{2.5ex} $a$ & $1.72$ & $d$ & $28.88$ & $\Tilde{a}$ & $1.64$ & $\Tilde{d}$ & $1.66$ \\ \hline
\rule{0pt}{2.5ex} $b$ & $13.32$& $e$ & $5.99$ & $\Tilde{b}$  & $0.33$ & $\Tilde{e}$  & $0.38$ \\ \hline
\rule{0pt}{2.5ex} $c$ & $0.29$ & $g$ & $-0.15$ & $\Tilde{c}$ & $0.02$ & $C_r$  &  $0.39$     \\ \hline
\rule{0pt}{2.5ex} $D$ & $2.98$ & $C$ & $0.69$ & $B$ & $0.29$ & $E$  &  $-3.07$     \\ \hline
\end{tabular}
\caption{Identified parameter values.}
\label{tab:fitting parameter values}
\end{table}

\section{Use cases}
This section presents an overview of work that featured DART as a test bed\footnote{The aim of this section is to showcase the platform's capabilities thus we will not provide an in-depth discussion on the scientific merits of the described experiments.}. From a functional point of view all these applications can be seen as a tracking problem, where a vehicle needs to follow a path at a certain reference speed. This reference speed may be fixed, or may be evaluated at runtime based on the robot's global position or on the position and/or velocity of other robots. To perform this planning task the robot needs a good dynamic model in order to track the reference velocity, access to its own pose (position and orientation) in the global reference frame, and access to the other robots' poses if needed. This can be achieved by means of an external motion capture system if it's available, or by means of on-board localization. The latter is however a viable option only for low to medium speeds, since the two key components are a sensor to perceive the environment, such as a camera or a lidar, and odometry data provided by the kinematic bicycle model (see section~\ref{sec:system identification}).
In the remainder of the section we will describe the main features of each kind of experiment, focusing on the functional requirements the platform needed to meet.

\subsubsection{Distributed MPCC~\cite{distributed_MPC}} 
\begin{figure}[t]
	\centering
 \setlength{\tabcolsep}{0.05em}
	  \begin{tabular}{cc}
    \includegraphics[width=.5\columnwidth]{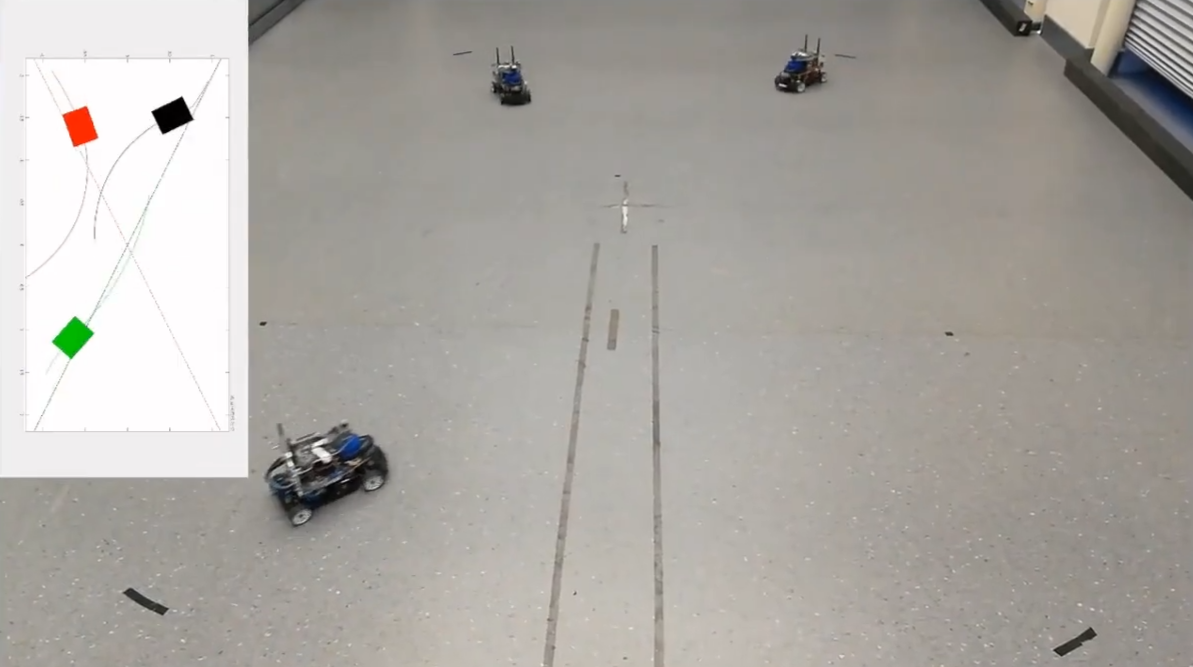} &
  \includegraphics[width=.5\columnwidth]{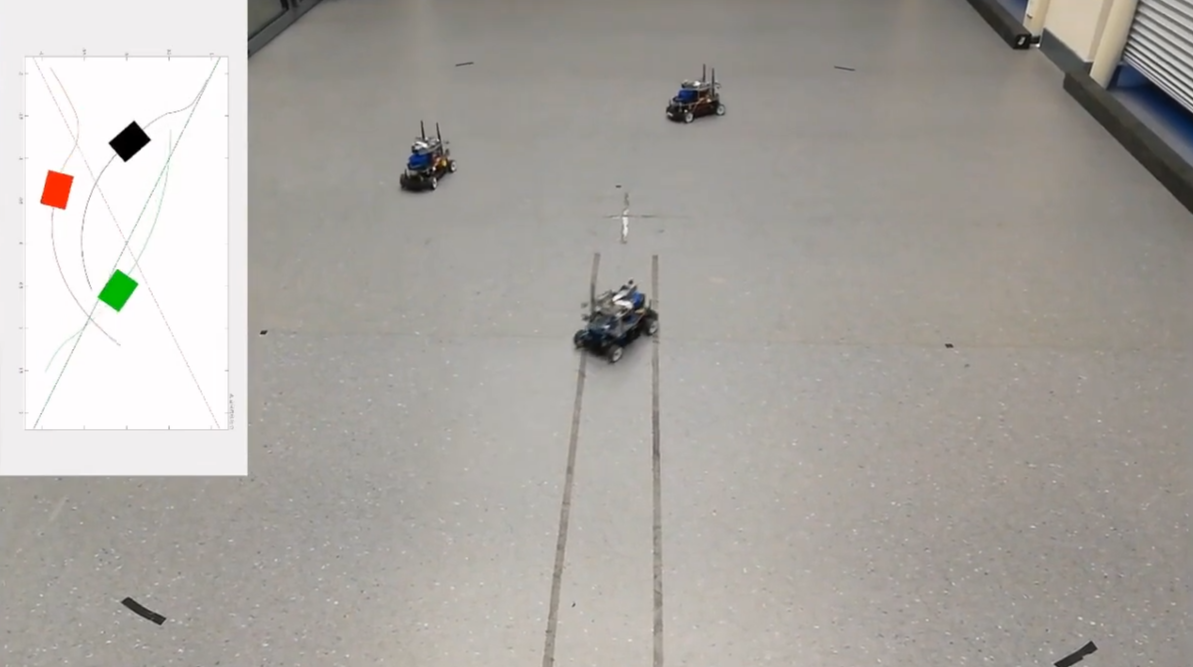} \\
  (a)&(b)\\
  \includegraphics[width=.5\columnwidth]{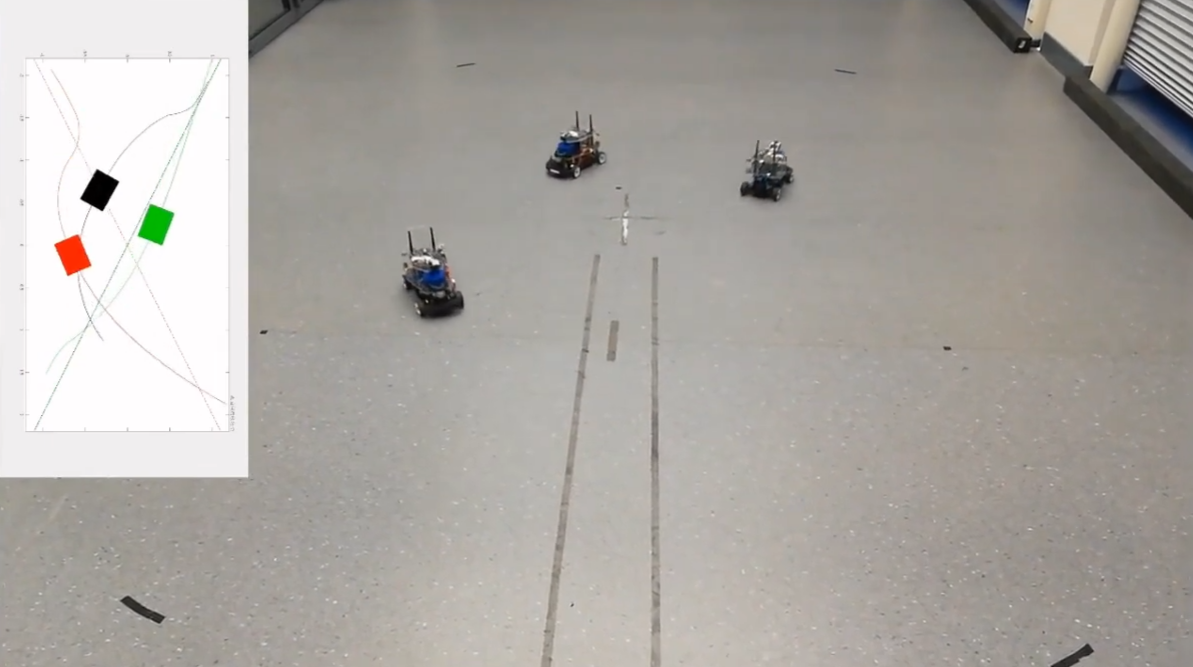} &
  \includegraphics[width=.5\columnwidth]{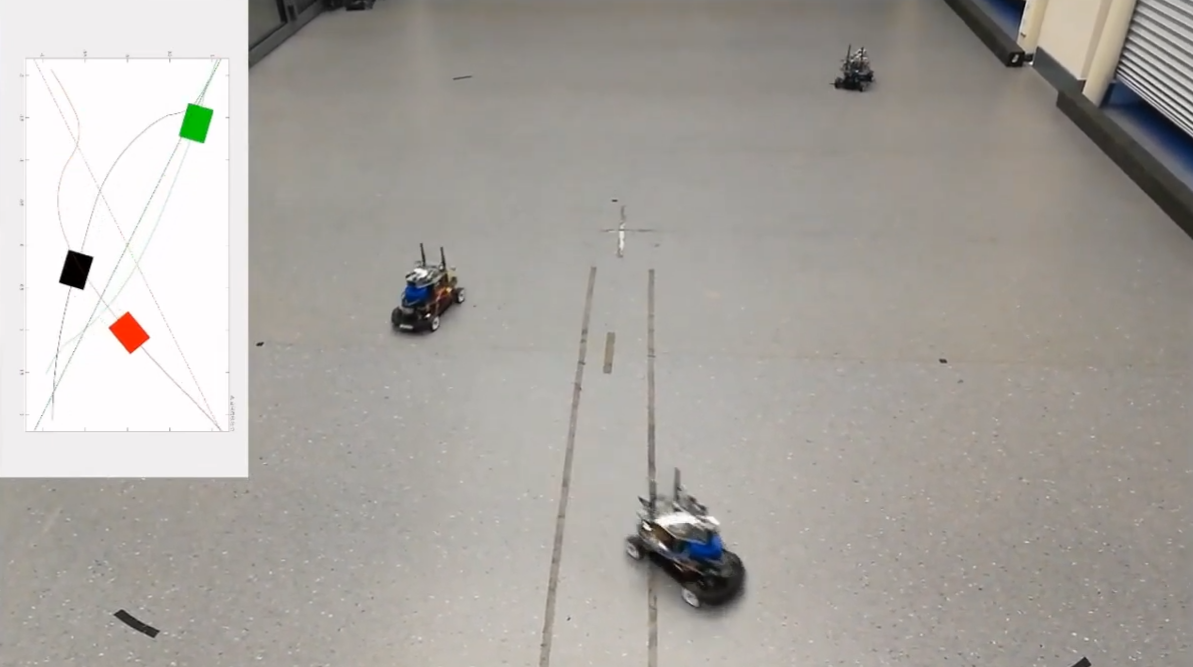} \\
  (c)&(d)\\
    \end{tabular}
	\caption{A team of robots navigate through an unsupervised intersection crossing using a distributed MPC scheme. This figure has been taken from~\cite{distributed_MPC}.}\label{fig:distributed mpc}
\end{figure}

We presented a distributed Model Predictive Contouring Control algorithm (D-MPCC) for a team of robots. Each robot aims at following a certain path at a given reference speed, while avoiding collisions with other agents. This is achieved through a distributed computation scheme that also accounts for possible packet loss over the communication network. The algorithm was tested in an intersection crossing, shown in Figure~\ref{fig:distributed mpc}, and a lane merging scenario. In both cases the robots need to be aware of each other's position and intended future trajectory in order to avoid collisions. To achieve this a good vehicle model is required, to ensure a consistent behaviour.

\subsubsection{Persistent monitoring~\cite{persistent_monitoring}} 
\begin{figure}[t]
  \centering
  \setlength{\tabcolsep}{0.05em}
  \begin{tabular}{cc}
    \includegraphics[width=.5\columnwidth]{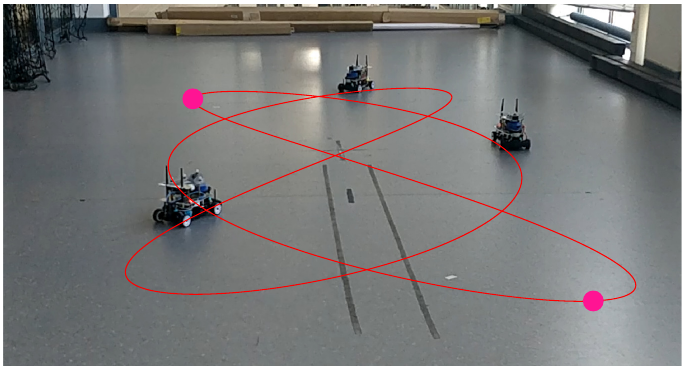} &
  \includegraphics[width=.5\columnwidth]{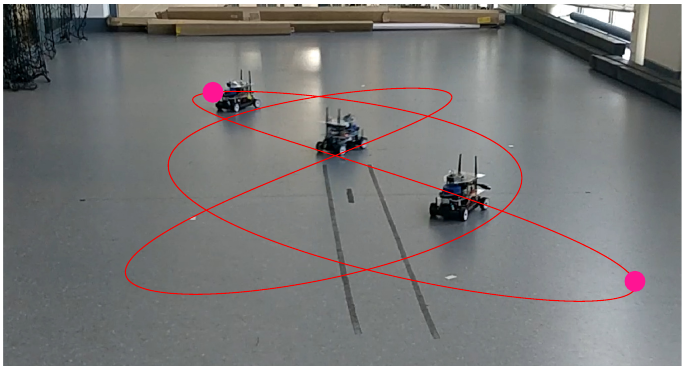} \\
  (a)&(b)\\
  \includegraphics[width=.5\columnwidth]{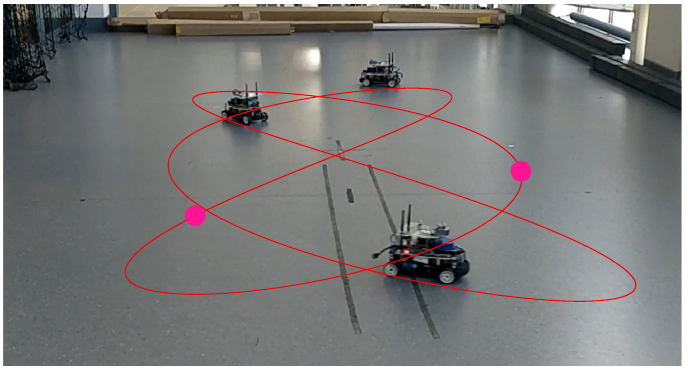} &
  \includegraphics[width=.5\columnwidth]{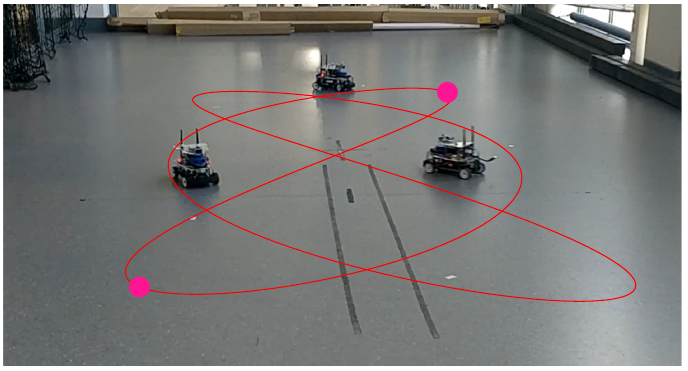} \\
  (c)&(d)\\
    \end{tabular}
    \caption{A team of robots following a Lissajous curve under a time-inverted Kuramoto dynamics feedback controller. This figure has been taken from~\cite{persistent_monitoring} .}
  \label{fig:kuramoto}
\end{figure}

A team of robots is tasked with monitoring a certain area. Each robot is equipped with omnidirectional sensors that are able to detect a target within a certain range. By relying on Lissajous curves and time-inverted Kuramoto dynamics, all vehicles follow the same smooth path within the designated area and adjust their speed based on the current position of the preceding and following vehicles, as shown in Figure~\ref{fig:kuramoto}. The emerging behaviour of the mobile sensor network is guaranteed to detect a moving target within bounded time and avoid collisions among agents. To successfully carry out the experiments the vehicles need to follow a highly curved path while accurately control their speed, since the latter needs to be adjusted according to the local path curvature and to the position of the other robots.

\subsubsection{Vehicle platooning} 
\begin{figure}[t]
	\centering
	\includegraphics[width=\linewidth]{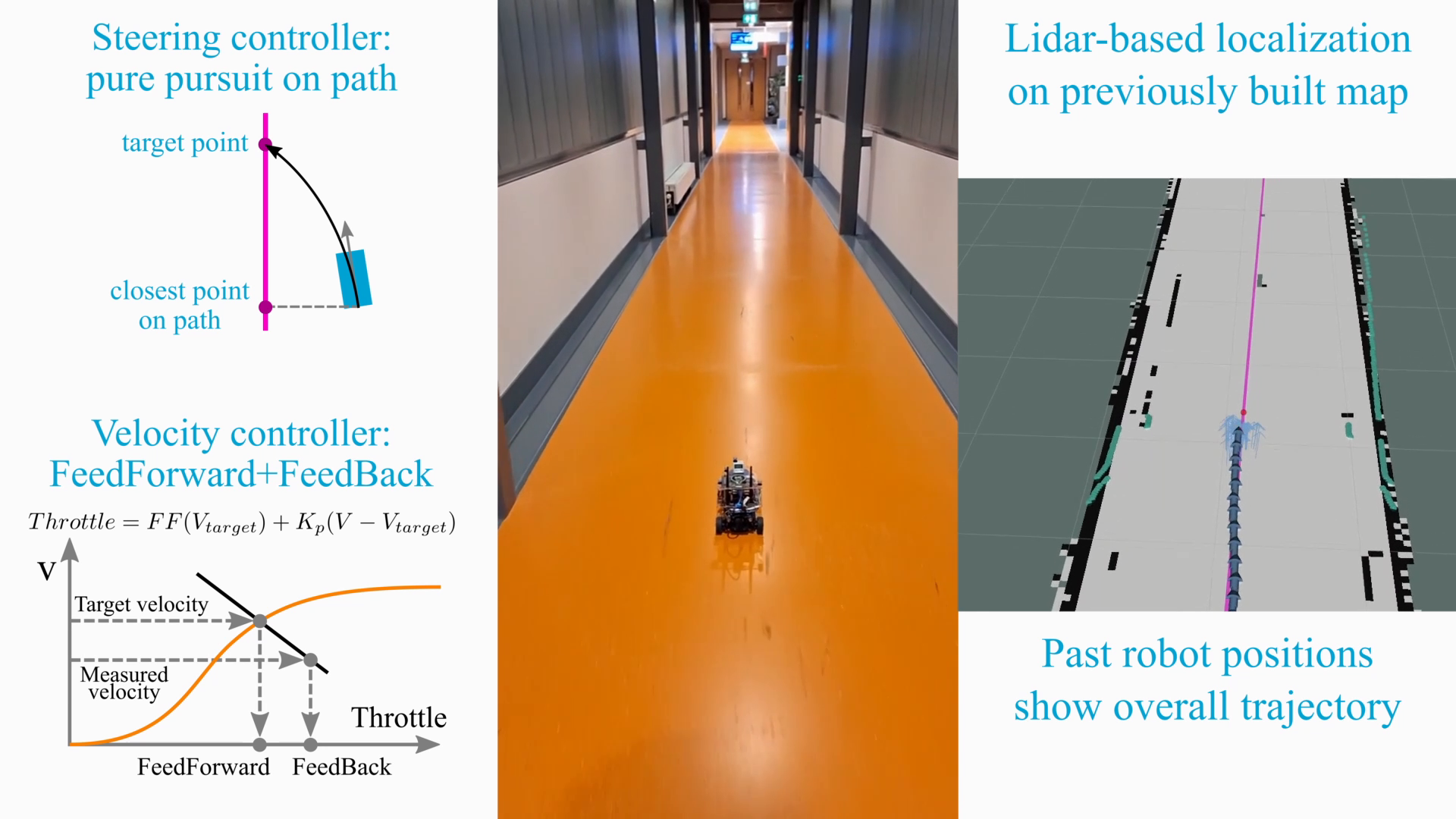}
	\caption{A robot following a straight line using on-board sensors for both localization and state feedback controllers. This image was taken from~\cite{lane_following_video}.}\label{fig:lane following}
\end{figure}

Platooning refers to the problem of vehicles driving along a relatively straight path while maintaining a certain distance among each other. The typical application is heavy duty trucks driving on a highway, where for small inter-vehicle distances significant fuel efficiency can be gained due to air drag reduction. To achieve this behaviour the vehicles need to share information on their current speed and position. From a practical point of view, the main challenge is that experiments require a long straight path, thus an external motion capture system will typically not be large enough, requiring the robots to rely on on board sensors for localization. Thanks to the lidar and on board odometry data this can be achieved using standard ROS libraries for Simultaneous Localization And Mapping (SLAM) and Adaptive Monte Carlo Localization (AMCL). Another significant challenge is to carefully adjust the steering in order to limit lateral deviations from the path. This is why we upgraded the servomotor used to steer the robot. Figure~\ref{fig:lane following} shows the robot using on-board localization and steering and velocity controllers to follow a straight path.

\subsubsection{Contouring MPC~\cite{CAMPCC}} 
\begin{figure}[t]
  \centering
  \setlength{\tabcolsep}{0.05em}
  \begin{tabular}{cc}
  \includegraphics[width=.5\columnwidth]{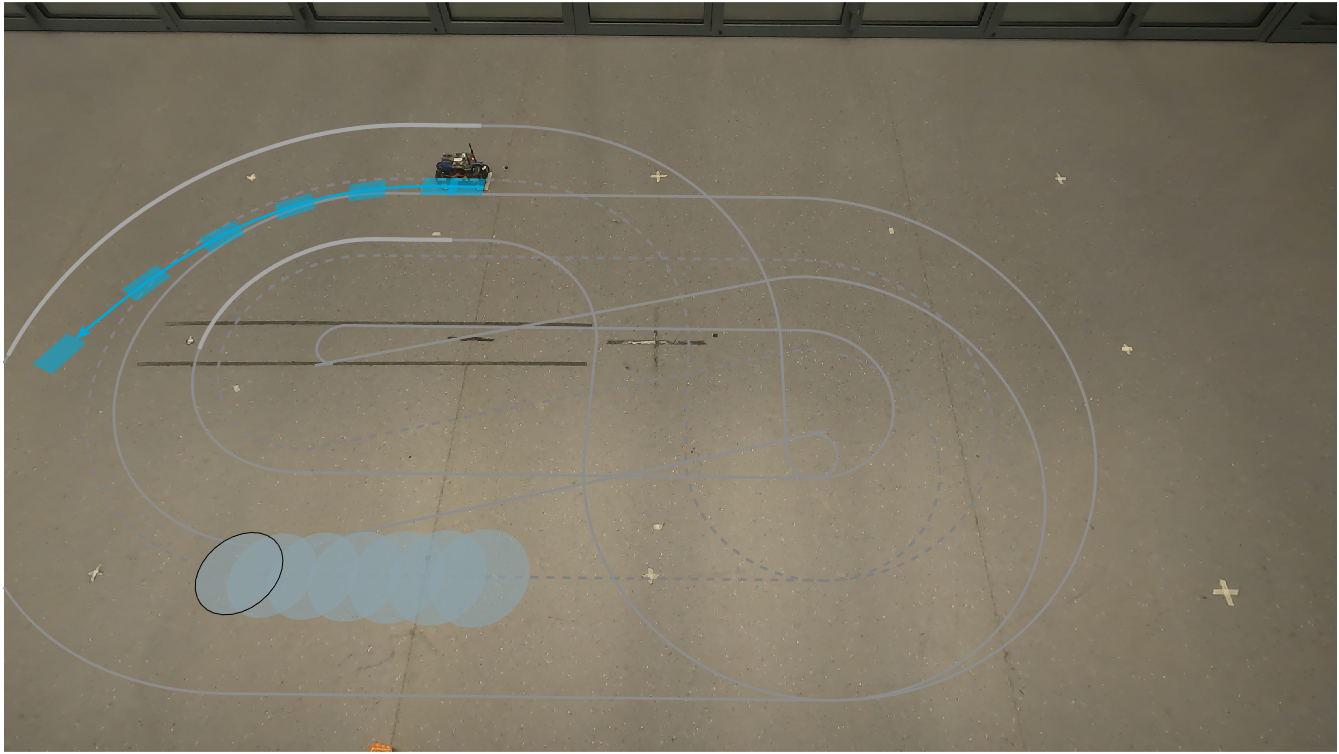} &
 \includegraphics[width=.5\columnwidth]{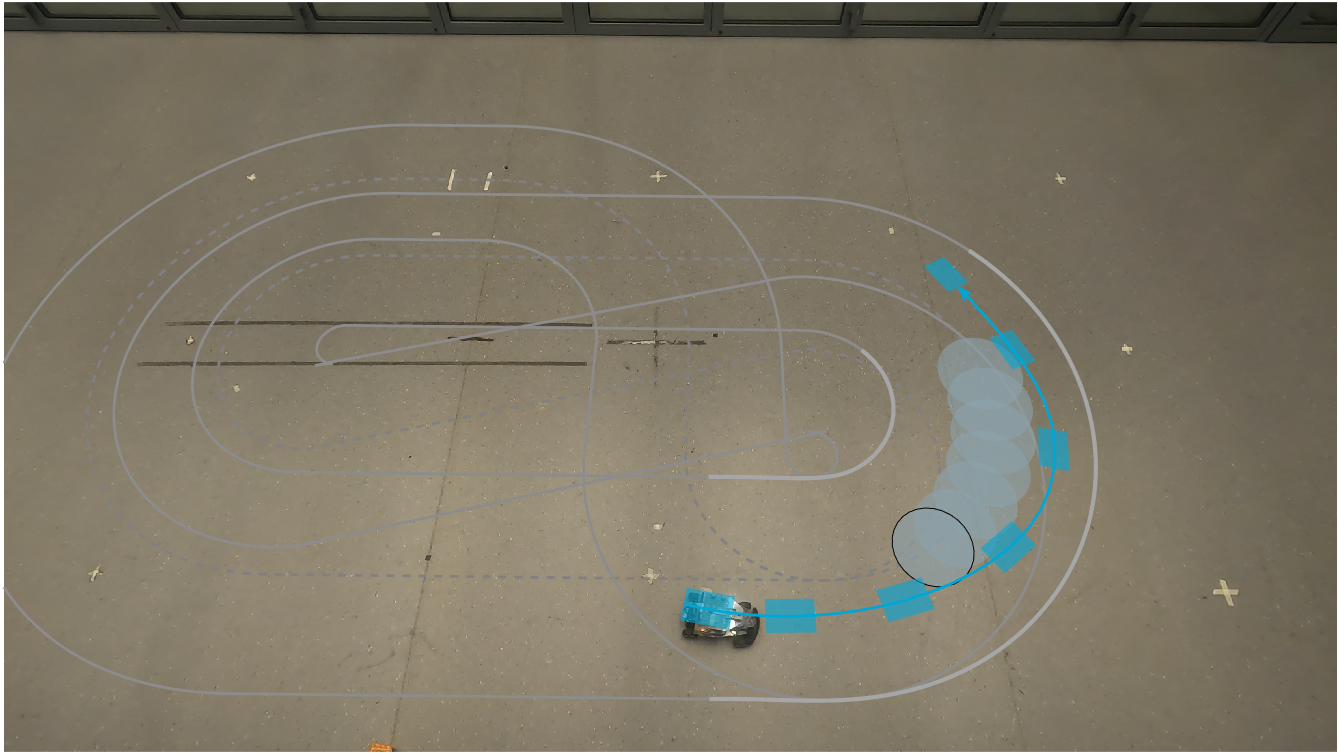} \\
   (a) $t=0$ s&(b) $t=8.6$ s\\
\includegraphics[width=.5\columnwidth]{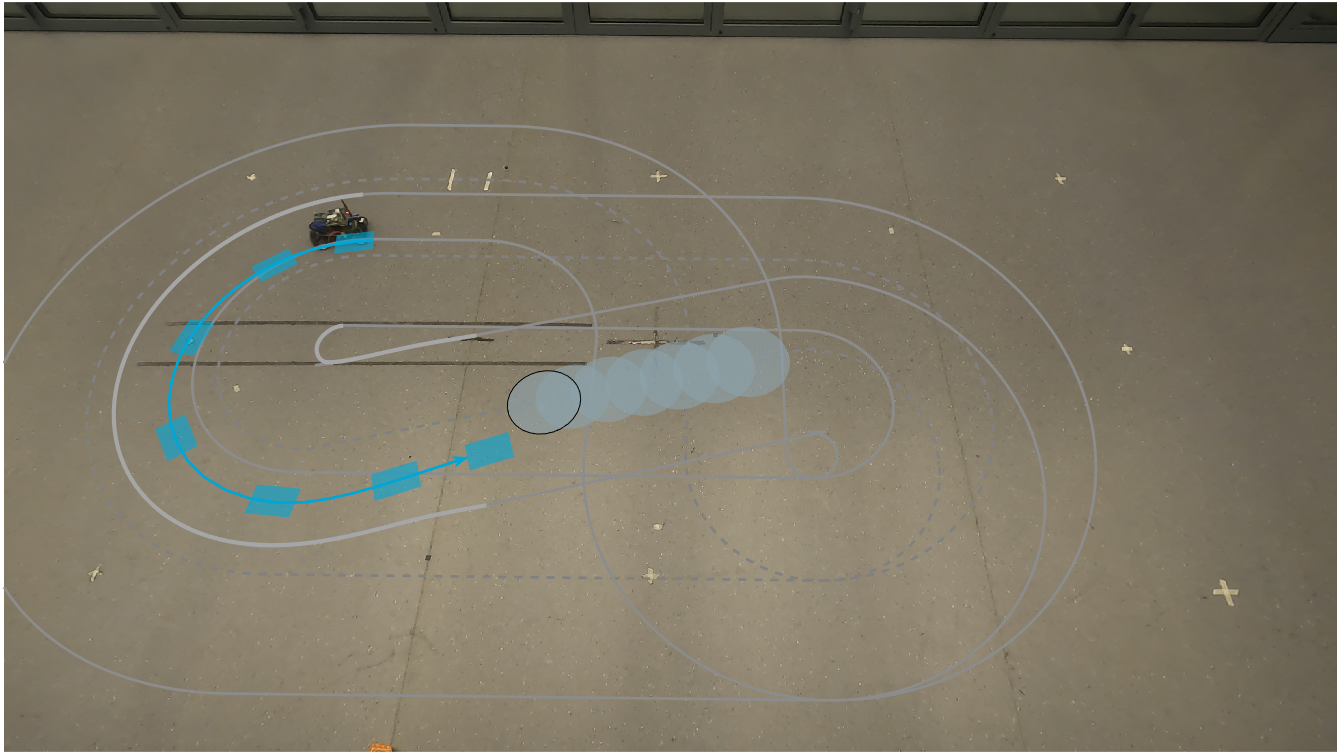} &
\includegraphics[width=.5\columnwidth]{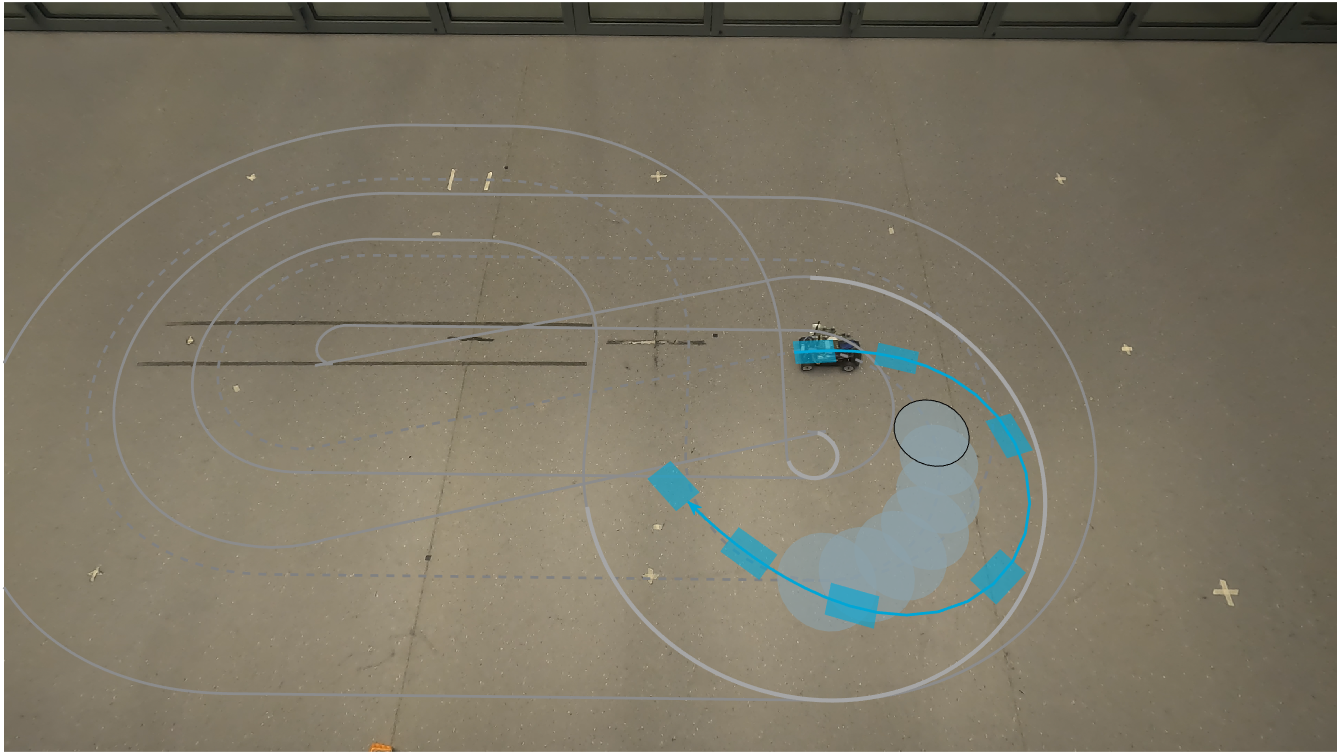}  \\
 (c) $t= 16.6$ s&(d) $t=22.4$ s\\
    \end{tabular}
    \caption{A robot following a highly curved path with the Curvature-Aware MPCC controller. This figure has been taken from~\cite{CAMPCC}.}
  \label{fig: CAMPCC experiments}
\end{figure}

We present a Model Predictive Contouring Control algorithm (MPCC) that also includes the information on the local path curvature, called Curvature-Aware MPCC. The new formulation features an improved estimation of the progress along the path and consequently more reliable lane boundary constraint satisfaction. Furthermore it features less cost function terms and is thus easier to tune. As far as experiments are concerned the main requirement for the platform is to exhibit consistent behaviour in order to highlight the differences due to the specific algorithm's formulation. This requires a good vehicle model. Figure~\ref{fig: CAMPCC experiments} shows the robot following a highly curved path while avoiding collision with a virtual dynamic obstacle.


\section{Conclusions}
This paper introduced the Delft's Autonomous-driving Robotic Test bed (DART), a small-scale car-like robot suitable for (multi-robot) motion planning and control. DART maximizes the use of available off-the-shelf hardware and features a low number of custom parts, making it easily reproducible and cost-effective. We provide a system identification procedure to obtain kinematic and dynamic bicycle models that allow the platform to be used for a wide range of applications, as demonstrated by the number of published works that featured DART as a test bed. Finally, we provide a GitHub repository containing building instructions, the data and code used for the identification, as well as a simulation environment and some readily-available low-level controllers.

\bibliographystyle{IEEEtran}
\bibliography{IEEEabrv,IV_small_scale_experimental_platform}

\begin{thebibliography}{10}
\providecommand{\url}[1]{#1}
\csname url@samestyle\endcsname
\providecommand{\newblock}{\relax}
\providecommand{\bibinfo}[2]{#2}
\providecommand{\BIBentrySTDinterwordspacing}{\spaceskip=0pt\relax}
\providecommand{\BIBentryALTinterwordstretchfactor}{4}
\providecommand{\BIBentryALTinterwordspacing}{\spaceskip=\fontdimen2\font plus
\BIBentryALTinterwordstretchfactor\fontdimen3\font minus \fontdimen4\font\relax}
\providecommand{\BIBforeignlanguage}[2]{{%
\expandafter\ifx\csname l@#1\endcsname\relax
\typeout{** WARNING: IEEEtran.bst: No hyphenation pattern has been}%
\typeout{** loaded for the language `#1'. Using the pattern for}%
\typeout{** the default language instead.}%
\else
\language=\csname l@#1\endcsname
\fi
#2}}
\providecommand{\BIBdecl}{\relax}
\BIBdecl

\bibitem{Web_article_robotaxis_worldwide}
\BIBentryALTinterwordspacing
E.~Juliussen, ``{Robotaxis: What Is Going On?}'' 8 2023. [Online]. Available: \url{https://www.eetimes.eu/robotaxis-what-is-going-on/}
\BIBentrySTDinterwordspacing

\bibitem{Web_article_paltooning_companies}
\BIBentryALTinterwordspacing
Y.~Rajan, ``{Top 5 truck platooning brands syncing convoy trucks via connective technology},'' 1 2022. [Online]. Available: \url{https://www.verifiedmarketresearch.com/blog/top-truck-platooning-brands/}
\BIBentrySTDinterwordspacing

\bibitem{Web_article_realistic_prospects}
\BIBentryALTinterwordspacing
N.~Winton, ``{Computer Driven Autos Still Years Away Despite Massive Investment},'' 2 2022. [Online]. Available: \url{https://www.forbes.com/sites/neilwinton/2022/02/27/computer-driven-autos-still-years-away-despite-massive-investment/}
\BIBentrySTDinterwordspacing

\bibitem{carla}
A.~Dosovitskiy, G.~Ros, F.~Codevilla, A.~Lopez, and V.~Koltun, ``{CARLA}: {An} open urban driving simulator,'' in \emph{Proceedings of the 1st Annual Conference on Robot Learning}, 2017, pp. 1--16.

\bibitem{khusro2020}
Y.~R. Khusro, Y.~Zheng, M.~Grottoli, and B.~Shyrokau, ``{MPC-based motion-cueing algorithm for a 6-DOF driving simulator with actuator constraints},'' \emph{Vehicles}, vol.~2, no.~4, pp. 625--647, 2020.

\bibitem{turtlebot}
\BIBentryALTinterwordspacing
ROBOTIS, ``{TurtleBot}.'' [Online]. Available: \url{https://www.turtlebot.com/}
\BIBentrySTDinterwordspacing

\bibitem{duckiebot}
\BIBentryALTinterwordspacing
Duckietown, ``{Duckiebot}.'' [Online]. Available: \url{https://get.duckietown.com/products/duckiebot-db21?variant=40700056895663}
\BIBentrySTDinterwordspacing

\bibitem{AUTORALLY}
B.~Goldfain, P.~Drews, C.~You, M.~Barulic, O.~Velev, P.~Tsiotras, and J.~M. Rehg, ``Autorally: An open platform for aggressive autonomous driving,'' \emph{Control Systems Magazine}, vol.~39, pp. 26--55, 2019.

\bibitem{MIT_RACECAR}
``\text{MIT RACECAR},'' \url{https://racecar.mit.edu/}.

\bibitem{BARC}
``\text{BARC},'' \url{https://closestnum20.com/barc_v4-0/}.

\bibitem{MUSHR}
\BIBentryALTinterwordspacing
S.~S. Srinivasa, P.~E. Lancaster, J.~Michalove, M.~Schmittle, C.~Summers, M.~Rockett, J.~R. Smith, S.~Choudhury, C.~Mavrogiannis, and F.~Sadeghi, ``Mushr: A low-cost, open-source robotic racecar for education and research,'' \emph{ArXiv}, vol. abs/1908.08031, 2019. [Online]. Available: \url{https://api.semanticscholar.org/CorpusID:201125218}
\BIBentrySTDinterwordspacing

\bibitem{DONKEY_CAR}
``\text{Dockey Car},'' \url{https://http://docs.donkeycar.com/}.

\bibitem{git_repo}
L.~Lorenzo, ``\text{DART},'' \url{https://github.com/Lorenzo-Lyons/DART}, 2024.

\bibitem{jetracer}
``\text{JetRacerProAI},'' \url{https://www.waveshare.com/wiki/JetRacer_Pro_AI_Kit}.

\bibitem{system_id}
L.~Ljung, ``Perspectives on system identification,'' \emph{Annual Reviews in Control}, vol.~34, no.~1, pp. 1--12, 2010.

\bibitem{rajamani2011vehicle}
R.~Rajamani, \emph{Vehicle dynamics and control}.\hskip 1em plus 0.5em minus 0.4em\relax Springer Science \& Business Media, 2011.

\bibitem{pacejka1992magic}
H.~B. Pacejka and E.~Bakker, ``The magic formula tyre model,'' \emph{Vehicle system dynamics}, vol.~21, no.~S1, pp. 1--18, 1992.

\bibitem{lateral_dynamics_practices}
G.~Baffet, A.~Charara, and D.~Lechner, ``Estimation of vehicle sideslip, tire force and wheel cornering stiffness,'' \emph{Control Engineering Practice}, vol.~17, no.~11, pp. 1255--1264, 2009.

\bibitem{kin_dyn_bicycle_models_borrelli}
J.~Kong, M.~Pfeiffer, G.~Schildbach, and F.~Borrelli, ``Kinematic and dynamic vehicle models for autonomous driving control design,'' in \emph{2015 IEEE intelligent vehicles symposium (IV)}.\hskip 1em plus 0.5em minus 0.4em\relax IEEE, 2015, pp. 1094--1099.

\bibitem{brushed_dc_motors}
``\text{Characteristics of Brushed DC Motors},'' \url{https://techweb.rohm.com/product/motor/brushed-motor/brushed-motor-basic/209/}.

\bibitem{cross_correlation}
\BIBentryALTinterwordspacing
W.~Menke, ``Chapter 9 - detecting and understanding correlations among data,'' in \emph{Environmental Data Analysis with MatLab® or Python (Third Edition)}, third edition~ed., W.~Menke, Ed.\hskip 1em plus 0.5em minus 0.4em\relax Academic Press, 2022, pp. 277--317. [Online]. Available: \url{https://www.sciencedirect.com/science/article/pii/B9780323955768000064}
\BIBentrySTDinterwordspacing

\bibitem{distributed_MPC}
L.~Ferranti, L.~Lyons, R.~R. Negenborn, T.~Keviczky, and J.~Alonso-Mora, ``Distributed nonlinear trajectory optimization for multi-robot motion planning,'' \emph{IEEE Transactions on Control Systems Technology}, vol.~31, no.~2, pp. 809--824, 2022.

\bibitem{persistent_monitoring}
M.~Boldrer, L.~Lyons, L.~Palopoli, D.~Fontanelli, and L.~Ferranti, ``Time-inverted kuramoto model meets lissajous curves: Multi-robot persistent monitoring and target detection,'' \emph{IEEE Robotics and Automation Letters}, vol.~8, no.~1, pp. 240--247, 2022.

\bibitem{CAMPCC}
L.~Lyons and L.~Ferranti, ``Curvature-aware model predictive contouring control,'' 2023.

\bibitem{lane_following_video}
\BIBentryALTinterwordspacing
{Reliable Robot Control lab}, ``{Lidar-based lane following}.'' [Online]. Available: \url{https://youtu.be/QgXHbPtGkdc}
\BIBentrySTDinterwordspacing

\end{thebibliography}

\end{document}